# A unified framework of predicting binary interestingness of images based on discriminant correlation analysis and multiple kernel learning

Qiang Sun[1], Liting Wang[1], Maohui Li[1], Longtao Zhang[1], Yuxiang Yang[2]

1. School of Automation and Informaiton Engineering, Xi'an University of Technology, Xi'an 710048, China
2. Department of Electronic Engineering, Hunan Normal University, Changsha 410081, China

**Abstract**
In the modern content-based image retrieval systems, there is an increasingly interest in constructing a computationally effective model to predict the interestingness of images since the measure of image interestingness could improve the human-centered search satisfaction and the user experience in different applications. Typically, a couple of interestingness cues are deliberately selected and the related features to capture these cues are elaborately leveraged in order to comprehensively represent the interestingness content underlying the images to be considered. How to make good use of these cues as well as features to arrive at an effective model to predict the interestingness of images is the core issue to address. In this paper, we propose a unified framework to predict the binary interestingness of images based on discriminant correlation analysis (DCA) and multiple kernel learning (MKL) techniques. More specially, on the one hand, to reduce feature redundancy in describing the interestingness cues of images, the DCA or multi-set discriminant correlation analysis (MDCA) technique is adopted to fuse multiple feature sets of the same type for individual cues by taking into account the class structure among the samples involved to describe the three classical interestingness cues, unusualness, aesthetics as well as general preferences, with three sets of compact and representative features; on the other hand, to make good use of the heterogeneity from the three sets of high-level features for describing the interestingness cues, the SimpleMKL method is employed to enhance the generalization ability of the built model for the task of the binary interestingness classification. Experimental results on the publicly-released interestingness prediction data set have demonstrated the rationality and effectiveness of the proposed framework in the binary prediction of image interestingness where we have conducted several groups of comparative studies across different interestingness feature combinations, different interestingness cues, as well as different feature types for the three interestingness cues.

# 1 Introduction

In the big data era, people have an increasingly strong interest in searching out the desired images with universally-recognized *interestingness* from different portal websites, which will become a higher requirement for the existing content-based image retrieval systems working in principle in terms of the relevance to a query. Therefore, the measure of interestingness of images will improve the human-centered search satisfaction and boost the user experience in different applications related to image retrieval functionality. Even though there is subjective difference in the judgement of the interestingness of an image, there is a substantial agreement on whether an image is interesting or not among observers. Thus, how to devise a computationally effective model to predict the interestingness of images becomes one critical issue, not only for image retrieval but also for other applications related to perception-level image quality assesment.

To implement the automatic prediction of the interestingness of an image, the first issue to address is to determine the related factors, or called cues, that visually attract human interest in images. For this problem, many scholars have made great efforts. Berlyne's research work on interestingness shows that interestingness is determined by some factors and their combinations, such as curiosity, uncertainty, conflict and complexity[1]. Chen et al. [2] verified through a lot of experiments that novelty, challenge, attention demand, intention and pleasure are the sources of interest. Chu et al. [3] studied the influence of familiarity on the perception of image interestingness. For example, unfamiliar background and familiar face images can make people feel interesting. Halonen et al. [4] also identified a series of other attributes related to visual interestingness, such as emotion, color, composition, genre and personal preferences. Gygli et al. [5] investigated the correlation of interestingness with an extensive list of image attributes based on the data set of specially designed and annotated images, finding that unusualness, aesthetics and general preferences are the three important and implemenTable cues in the description of image interestingness across a variety of contexts, which is the basis on which we develop our interestingness prediction scheme.

The second issue is then to determine the relevant features that can capture individual cues of interestingness in images. In a series of research schemes released in 2016 to predict multimedia interesting tasks[6-10], the organizer provided the participants with a set of descriptors (features): scale invariant feature transformation (SIFT), local binary pattern (LBP), histogram of oriented gradient (HOG), color histogram in the HSV space, Gist, and deep learning-based features. Based on the same data set, participants used different features and combinations to build reasonable computational models to predict the interestingness of images. In the work of [5], to capture each of the three cues above for interestingness, i.e. unusualness, aesthetics and general preferences, Gygli et al. explored and chose various features for these cues in varying contexts. On the whole, these features have more or less efficacy in predicting the interestingness of images, depending on their content and the contexts. Due to the difference of variability and complexity among images, more features could be utilized to extend the list of features to comprehensively describe different cues affecting the interestingness decision.

The third issue to deal with is how to make good use of the extracted features for different interestingness cues. In [5], starting from the best single feature for describing the interestingness of the image involved, Gygli et al. performed greedy forward feature selection to combine the individual features they have proposed for the three cues of unusualness, aesthetics and general preferences. This combination of features is not one principled way to utilize the features since the best feature is not uniform for every image and depends upon the image. Moreover, greedy feature selection by gradual selection of additional features is extremely time-consuming, especially in the case where a large number of images need be involved to facilitate the training process. In fact, for each feature type that reflects different interestingness cues, a few related features need to be extracted to fully represent the characteristics of the type. As such, feature redundancy will be induced inevitably, which will become a key problem to solve in this procedure.

The final issue is to make decision about the interestingness/uninterestingness of test images. In this procedure, one classifer is typically employed to implement the binary classification task of image interestingness. As a machine learning method, support vector machine (SVM) could be used to perform binary classification task, where the feature space is divided to find the optimal hyperplane among samples of different classes to complete the linear classification of samples. More importantly, the kernel function of SVM enables nonlinear data to become linearly separable through mapping the original data to the high-dimensional space, thus the optimal classification hyperplane can be constructed in the new space to complete the standard linear classification task. One can use the kernel function of SVM to evaluate the similarity between two vectors, where different kernels are derived from different information sources. In practice, most data are heterogeneous, like the features extracted for different interestingness cues. Compared with a single kernel, different kernel combinations make the use of kernel methods more flexible and more interpreTable, which is the motivation of the Multiple Kernel Learning (MKL) technique[12]. Therefore, the robustness of the interestingness classification should be expected to boost with the MKL method. To the best of our knowledge, there is no similar work in the field of image interestingness understanding.

To implement the binary interestingness classification of images, this paper makes four important contributions:

i) We extend, to some degree, the features for the three cues of unusualness, aesthetics and general preferences, which is beneficial to comprehensively describe the factors that may reflect the interestingness of images.

ii) In order to solve the problem of feature redundancy in capturing the cues of interestingness of images, this paper adopts the discriminant correlation analysis (DCA) or multiset discriminant correlation analysis (MDCA) technique in [13] to fuse multiple features of the same type for individual cues by taking into account the class structure among samples, and finally describes the three cues above with three sets of compact and representative features.

iii) In order to make full use of the heterogeneity from the three sets of high-level features for different interestingness cues, the simple multiple kernel learning (SimpleMKL) [14] method is employed to enhance the generalization ability of the built model for the task of the binary interestingness classification.

iv) In terms of quantitative metrics, several groups of comparative studies across different interestingness feature combinations, different interestingness cues, as well as different feature types for the three interestingness cues are carried out to examine how important these cases are to predict the interestingness of images.

This paper is organized as follows: The related works, i.e. DCA technique and the SimpleMKL algorithm, are briefly introduced in Section II. Section III describes our proposed framework for binary interestingness classification and the technical details in different procedures. The experimental settings, results and detailed analysis on the public database are presented in Section IV. Finally, Section V concludes the paper.

## 2  Related Work

In this section, we provide the brief overview of the DCA technique [13] and the SimpleMKL algrithm[14], each of which will be leveraged to address the feature fusion problem and the MKL-based classification problem respectively.

### 2.1 Discriminant Correlation Analysis

Compared to the CCA, the DCA, proposed in [13], not only maximizes the correlation of corresponding features between two feature sets, but also separates the classes within each feature set[15]. To this end, a two-step method is designed for the DCA: 1) the separate discriminant analysis for each feature set with the inter-class scattering matrices, and 2) the correlation analysis between feature sets driven by the diagonalization of the between-set covariance matrix. Both steps of the DCA will be overviewed as follows.

Let the matrices of $X \in R^{p \times n}$ and $Y \in R^{q \times n}$ denote two different feature sets respectively, each consisting of $n$ $p$-dimensional, or $q$-dimensional, feature vectors extracted from different modalities. The $n$ columns in the two matrices are composed of $c$ different groups, each group associated to one of the $c$ classes, and $n = \sum_{i=1}^{c} n_i$ where $n_i$ columns belong to the $i^{th}$ class. Let $x_{ij} \in X$ denote the $j^{th}$ feature vector for the $i^{th}$ class, and $\bar{x}_i$ and $\bar{x}$ be the mean of the $i^{th}$ class and that of the whole feature sets respectively. Let $S_{bx}$ denote the inter-class scattering matrix of the feature set $X$ as

$$S_{bx_{(p \times p)}} = \sum_{i=1}^{c} n_i (\bar{x}_i - \bar{x})(\bar{x}_i - \bar{x})^T = \Phi_{bx} \Phi_{bx}^T \qquad (1)$$

where $\Phi_{bx_{(p \times c)}} = [\sqrt{n_1}(\bar{x}_1 - \bar{x}), \sqrt{n_2}(\bar{x}_2 - \bar{x}), ..., \sqrt{n_c}(\bar{x}_c - \bar{x})]$.

The first step of DCA is to find one transformation matrix $W_{bx}$ that can project the feature matrix $X_{(p \times n)}$ onto one $r$-dimensional feature space to get the new feature matrix $X'_{(r\times n)} = W^T_{bx_{(r\times n)}} X_{p\times n}$. Accordingly, the inter-class scattering matrix $S_{bx}$ can be updated as $S'_{bx} = W^T_{bx} S_{bx} W_{bx} = \Phi'^T_{bx} \Phi'_{bx} = I$. The scattering matrix $\Phi'^T_{bx} \Phi'_{bx}$ is a strictly diagonally dominant matrix where the diagonal elements are close to one and the non-diagonal elements close to zero, implying that the centroids of the classes have minimal correlation with each other, and thus the classes are separated in the projected space. In other words, the updated features derived from this transformation are characterized by the class discriminability. About the way how to the obtain the transformation matrix $W_{bx}$, please refer to the details in [13].

Similarly, one can use DCA to solve for the second feature set $Y_{(q\times n)}$ to find the corresponding transformation matrix $W_{by}$ to unitize the inter-class scattering matrix for the second modality, $S_{by}$, to obtain the reduced-dimension feature set $Y'_{(r\times n)} = W^T_{by_{(r\times q)}} Y_{q\times n}$. The updated inter-class scattering matrix is then $S'_{by} = W^T_{by} S_{by} W_{by} = \Phi'^T_{by} \Phi'_{by} = I$, which also means that the classes for the second modality are separated in the new space.

Taking into account the fact that there are at most $c-1$ nonzero generalized eigenvalues and that other upper bounds for $r$ are the ranks of the feature matrices of $X$ and $Y$, it can be easily inferred that $r \leq \min(c-1, rank(X), rank(Y))$ [13].

The second step is to diagonalize the between-set covariance matrix $S'_{xy} = X'Y'^T$ of the transformed feature sets, $X'$ and $Y'$, making the features in the feature set of $X$ have non-zero correlation only with the corresponding features and zero correlation with the non-corresponding features in the feature set of $Y$. To this end, the singular value decomposition (SVD) is employed to diagonalize $S'_{xy}$ as follows:

$$S'_{xy_{(r\times r)}} = U\Sigma V^T \Rightarrow U^T S'_{xy} V = \Sigma \tag{2}$$

where $\Sigma$ is a diagonal matrix with non-zero elements on the primary diagonal. Let $W_{cx} = U\Sigma^{-\frac{1}{2}}$ and $W_{cy} = V\Sigma^{-\frac{1}{2}}$, one can have

$$(U\Sigma^{-\frac{1}{2}})^T S'_{xy} (V\Sigma^{-\frac{1}{2}}) = I \Rightarrow W^T_{cx} S'_{xy} W_{cy} = I \tag{3}$$

Thus, $W_{cx}$ and $W_{cy}$ are the transformation matrix for $X'$ and $Y'$ and the resulting transformed feature sets are written as

$$\hat{X} = W^T_{cx} X' = W^T_{cx} W^T_{bx} X \tag{4}$$

$$\hat{Y} = W_{cy}^T Y' = W_{cy}^T W_{by}^T Y$$

The inter-class scattering matrix $\hat{S}_{bx}$ of $\hat{X}$ can be written as

$$\hat{S}_{bx} = W_{cx}^T W_{bx}^T S_{bx} W_{bx} W_{cx} \tag{5}$$

Taking into consideration of the fact that $W_{bx}^T S_{bx} W_{bx} = I$ and that $U$ is an orthogonal matrix, the following equation holds:

$$\hat{S}_{bx} = \left(U\Sigma^{-\frac{1}{2}}\right)^T \left(U\Sigma^{-\frac{1}{2}}\right) = \Sigma^{-1} \tag{6}$$

which indicates that the inter-class scattering matrix $\hat{S}_{bx}$ is still diagonal, and thus the classes remain separated. Similarly, one can infer that $\hat{S}_{by} = \Sigma^{-1}$, showing that $\hat{S}_{by}$ is also diagonal and the classes well-separated for the second feature set. As a result, it can be satisfied simultaneously that both the classes are separated for each feature set and the correlation between the corresponding features and the uncorrelation between the non-corresponding features across two feature sets are maximized.

## 2.2 Simple Multiple Kernel Learning (SimpleMKL)

Generally speaking, the solution of the kernel-based learning method, like the well-known SVM, has the following form as

$$f(\mathbf{x}) = \sum_{i=1}^{l} \alpha_i^* K(\mathbf{x}, \mathbf{x}_i) + b^* \tag{7}$$

where $\alpha_i^*$ and $b^*$ are the optimal coefficients to learned from examples, $l$ is the number of support vectors, $K(\mathbf{x}, \mathbf{x}_i)$ is a give positive-definite kernel, and $f(\mathbf{x})$ can be solved by a convex quadratic optimization algorithm.

In a wide range of practical applications, mutiple heterogeneous data sources are typically involved in the machine learning problems, which shows that it is benefical to enhance the flexibility of the learnt models with multiple kernels instead of one single fixed kernel, i.e., the multiple kernel learning (MKL) problem.

In principle, with $M$ kernels $K_m(\mathbf{x}, \mathbf{y})$, $m = 1, ..., M$, the common object in different MKL algorithms is to learn an optimal convex combination of these basic kernels as follows

$$K(\mathbf{x},\mathbf{y}) = \sum_{m=1}^{M} d_m K_m(\mathbf{x},\mathbf{y}), \quad \text{s.t.} \sum_{m=1}^{M} d_m = 1, d_m \geq 0 \; \forall m \tag{8}$$

where $d_m$ is the weight of the $m$th kernel. In other words, learning both the coefficients $\alpha_i^*$ and the weights $d_m$ in a single optimization problem is referred to as the MKL problem.

For the SimpleMKL algorithm in [14], the optimization problem to be solved has the following formulation:

$$\begin{cases} \min_{d} J(\mathbf{d}) \\ \text{s.t.} \sum_{m=1}^{M} d_m = 1, d_m \geq 0 \; \forall m \end{cases} \tag{9}$$

where the objective function $J(\mathbf{d})$ is defined as

$$J(\mathbf{d}) = \begin{cases} \min_{\{f\},b,\xi} \left( \dfrac{1}{2} \sum_{m=1}^{M} \dfrac{1}{d_m} \|f_m\|_{\mathcal{H}_m}^2 + C \sum_{i=1}^{N} \xi_i \right) \\ \text{s.t.} \; y_i \left( \sum_{m=1}^{M} f_m(x_i) + b \right) \geq 1 - \xi_i, \forall i \\ \xi_i \geq 0, \forall i \end{cases} \tag{10}$$

where each function $f_m$ belongs to a different reproducing kernel Hilbert space (RKHS) $\mathcal{H}_m$ associated with a given positive definite kernel $K_m(\mathbf{x},\mathbf{y})$. In the MKL framework, what one needs to do is to find a decision function of the form as $f(x) + b = \sum_m f_m(x) + b$.

According to the optimization theory, the dual problem of the above formulation can be written as

$$\begin{cases} \max_{\alpha} \left( -\dfrac{1}{2} \sum_{i=1}^{N} \sum_{j=1}^{N} \alpha_i \alpha_j y_i y_j \sum_{m=1}^{M} d_m k_m(x_i, x_j) + \sum_{i=1}^{N} \alpha_i \right) \\ \text{s.t.} \sum_{i=1}^{N} \alpha_i y_i = 0, 0 \leq \alpha_i \leq C, i = 1,2,\ldots,N \end{cases} \tag{11}$$

With the weights available, the above optimization problem becomes a classical SVM dual formulation.

Therefore, the objective function $J(\mathbf{d})$ can be rewritten as

$$J(\mathbf{d}) = -\frac{1}{2}\sum_{i=1}^{N}\sum_{j=1}^{N}\alpha_i^* \alpha_j^* y_i y_j \sum_{m=1}^{M} d_m K_m(\mathbf{x}_i, \mathbf{x}_j) + \sum_{i=1}^{N}\alpha_i^* \quad (12)$$

where $\alpha^*$ is the optimal solution of the formulation in Eq. (11), and $J(\mathbf{d})$ can be can be evaluated by any SVM algorithm. For more details on the SimpleMKL algorithm, please refer to [14].

Compared with its counterparts, the SimpleMKL algorithm is much more efficient for large-scale classification problems involved with a large number of data samples and multiple kernels [16]. The aim that we adopt it is to integrate various heterogeneous features across three interestingness cues extracted from indoor or outdoor images, and to enhance the classification performance.

## 3 Binary prediction method of image interestingness

### 3.1 Overview

The research framework in this paper is shown in Fig.1. It consists of four parts: input data (training images/test images), the extraction of relevant features for capturing the interestingness clues, feature fusion based on the DCA/MDCA technique[13], and the binary interestingness classification by the SimpleMKL algorithm[14].

The first part is the input data, for which we have used the data set provided in the competition of Predicting Multimedia Interestingness Task released in 2016 [17], in which the training images and test images are derived from the Hollywood movie trailers under the Creative Commons licenses. For the second part, three influential cues found in [5], i.e. unusualness, aesthetics and general preferences, are exploited to describe different aspects of the interestingness of an image. More specially, for the unusualness cue, two types of features, namely familiarity measure and local outlier coefficient, have been selected; for the aesthetic cue, five types of features including arousal, color, texture, complexity and shape were selected; for the cue of general preferences, we have used SIFT, HOG and Gist features. The third part is the fusion of features belonging to the same type using the DCA or MDCA technique. Then, different types of features are concatenated to form three feature sets that are related to the three interestingness cues above. The fourth part is the binary interestingness classification with the multiple kernel learning methodology. For different cues, distinct kernel functions are adopted, and thus the new multi-kernel function is composed of three kernel functions. In the training phase, the classical SimpleMKL algorithm [14] has been exploited to get the trained model, which was then used to predict the interestingness/uninterestingness of the test image in the test stage.

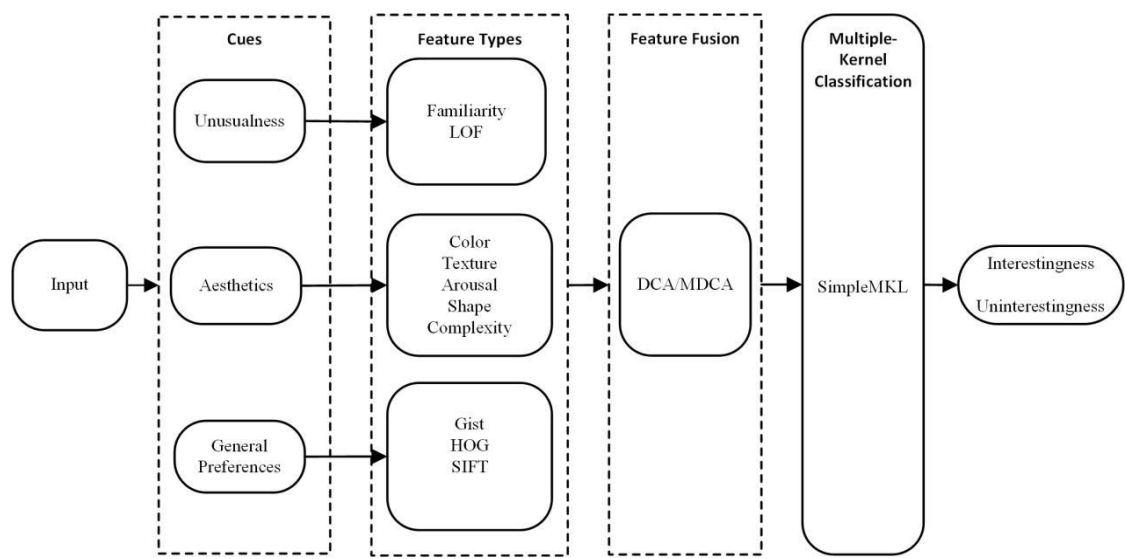

Fig. 1 The schematic of the proposed binary interestingness prediction framework for images

**3.2 Cues and features for the description of interestingness of images**

Since the seminal work by Berlyne[1] for exploring the core ingredients that affect the interestingness in images, many scholars have been making effort to explore the dominant aspects of interestingness from different images, indoor or outdoor ones. Among these works, Gygli et al. [5] have found three main cues, i.e. unusualness, aesthetics and general preferences, with high influence on the prediction of interesingness of an image, which is the basis on which we developed our interesingness prediction framework. More specially, part of the features in [5] has been adopted and other relevant features in the existing literature that could computationally capture these three cues have also been leveraged, more or less, to extend the former, which will be described briefly as follows.

**(1) Unusualness**

Unusualness, also called novelty, is one of the important and implementable cues for the prediction of interestingness [5]. In our work, the *Familiarity* feature [18] and Local Outlier Factor (LOF) feature [19] have also been extracted from each image to represent the unusualness cue.

1) *Familiarity*

The familiarity feature, proposed in [18], was calculated as the average distance between an image and its *k*-closest images in terms of local features. The further the distance, the more unusual it is [5].

2) *Local Outlier Factor*(LOF)

LOF features are extracted mainly by applying the local outlier factor algorithm [19] to different global descriptors. LOF is a density-based outlier measurement method. Specially, if the images within the *k* nearest neighbors of an image are relatively dense, the central image is normal, otherwise it is outlier [19]. In this paper, according to the similar approach in [5], we have also used the 10-distance neighborhood to calculate its local outlier coefficient, through which the outlier degree of an image was measured.

**(2) Aesthetics**

The studies show that there is a strong correlation between the interestingness and the aesthetics of an image [5]. Therefore, the research results of computational aesthetics in the exitsting literature could lay a solid foundation for the description of the aesthetics cue to predict the interestingness of images in our work. In the following, the feature types that have been utilized to capture the aesthetics cue are introduced, with each type being composed of a couple of features.

1) *Color*

In [5], the Earth Mover Distance (EMD) between the image color histograms in the LUV color space, originally proposed in [18], was adopted to describe the color information of the image as the color feature. In addition, we have also used color correlogram features [20] and three color moments [21] to describe the color features. The former contains not only the color statistics but also the spatial distribution information of colors with different distance transformations (i.e. the spatial relationship between colors), while the latter extracts as features the global statistics of the three color moments in the HSV color space and the mean of every moment.

2) *Texture*

Texture can be used to describe the influence of the brightness distribution and variation of an image on subjective perception, so it is an important feature to describe the aesthetics cue.

In our work, we extracted the texture features in three ways: i) Calculate the spatial texture features including Contrast, Energy, Entropy, Inverse Differential Moment and Correlation based on the GLCM [22] with four directions of 0 ,45 ,90 ,135 and the distance of 1; ii) Texture in an image owns multi-scale characteristics, and different details are distributed in different scales of the image. In view of this, we used Haar wavelet transform to obtain subband images in different scales and directions to represent the texture characteristics of an image; iii) Extract the Rotation Invariant Uniform Local Binary Patterns [23] as an effective supplement to the GLCM-based spatial texture features.

3) *Arousal*

According to the theory of arousal, pleasure and dominance space, color in an image can stir up a wide range of human emotions [24]. Based on this, Machadjik and Hanbury [25] extracted various emotional features from the raw pixels of an image. In our work, we followed the practice of [5] to capture the aesthetics of an image by combining the brightness and saturation of the image to evaluate the arousal score as feature since this emotion dimension is correlated with interestingness.

4) *Shape*

Shape reflects the information of the overall envelope of an image, which can be deployed to capture the aesthetics of the image from a global perspective [26]. Generally, the description methods of shapes can be divided into two types: the contour-based methods and the region-based ones. The former focuses on the outer contour of shapes, while the latter pays attention to the region information associated to the shape. In this paper, we extracted the edge histogram information using the Sobel operator and the HU invariant moments via the Canny edge detection operator to describe the shape information.

5) *Complexity*

The existing studies show that there is a close correlation between the complexity of and the aesthetics of an image [27] . The complexity of an image can be assessed according to different theoretical studies. In this paper, we extracted the complexity feature of images based on the information theory and image compression theory respectively. For the first case, we evaluated the entropy of an image as its information theory-based complexity feature. The second complexity feature is extracted based on the image compression theory where the complexity is measured in terms of the compression rate. The comparative study by Yu et al. [27] found that the mean of the Spatial Information (SI) image is an effective index on the complexity of an image. Silva et al. [28] pointed out that compared to the JPEG compression rate of the original image, the JPEG compression rate derived from the heat map or the saliency map of the image could represent the image complexity in a manner which is more consistent with the human visual perception mechanism. Machado et al. [29] utilized the image compression error and the Zipf's law to describe the visual complexity of an image, showing that the edge maps resulting from some edge detection technique could greatly enhance the correlation between the complexity features and the subjective experiences of human beings. Based on the above findings, the JPEG compression rate of the saliency maps generated by the salient object detection method in [30], the mean and the root mean square of the SI images, and the mean, the standard deviation and the compression rate of the edge maps by the Canny operator have been evaluated in our work as the complexity feature of an image.

**(3) General preferences**

Studies have proven that although people have some subjectivity to evaluate the interestingness of images, there are similarities among different subjective experiences. More specially, people typically have more interest in some scenes than others. In other words, they have general preferences for certain scenes [5]. In our work, we mainly used the global scene descriptors based on Gist [26], histograms of oriented gradient (HOG) [31] and spatial pyramids of scale invariant feature transform (SIFT) histograms [32] to capture this type of interestingness characteristics.

**3.3 Feature fusion based on the discriminant correlation analysis**

In our work, the discriminant correlation analysis (DCA) technique [13] was employed for the feature-level fusion. DCA is a significant improvement of the canonical correlation analysis (CCA) technique [33] which is an important statistical method to measure the multivariate correlation between two sets of variables. When the CCA is used to implement feature fusion, the two sets of transformations are needed to be found by calculating the correlation between two sets of features such that the transformed features have the maximum correlation between the two feature sets, and within each feature set there is no correlation or the correlation is minimum, which could produce the final canonical correlation features as the fused features [34]. However, the CCA-based fusion methods typically ignore the class structure among samples in different pattern recognition applications. To address this issue, the DCA not only inherits the advantage of the CCA in maximizing the correlation between the two feature sets, but also decorrelates the features that belong to different classes within each feature set [13], thus, during the feature fusion process, the class structures are fully taken into consideration to enhance the discrimination ability of feature fusion.

Based on the overview of the DCA in Section 2.1, let $W_x = W_{cx}^T W_{bx}^T$ and $W_y = W_{cy}^T W_{by}^T$ denote the final tranformation matrices for $X$ and $Y$, respectively. According to the classical fusion strategies [34], the feature-level fusion can be implemented either by the concatenation ($Z_1$) or the summation ($Z_2$) of the transformed feature vectors as follows:

$$Z_1 = \begin{pmatrix} \hat{X} \\ \hat{Y} \end{pmatrix} = \begin{pmatrix} W_x X \\ W_y Y \end{pmatrix} \tag{13}$$

$$Z_2 = \hat{X} + \hat{Y} = W_x X + W_y Y \tag{14}$$

where $Z_1$ and $Z_2$ are termed the Discriminant Correlation Features (DCFs). In our work, we adopted the concatenation scheme since there is, although slight, advantage for the results over those from the summation one.

The DCA-based feature fusion method has demonstrated its excellent performance in the application of multi-modal biometric recognition, but the remaining limitation is it is only applicable to the fusion between two feature sets rather than more. To solve this problem, Haghighat et al. [13] then generalized the DCA to multi-set Discriminant Correlation Analysis (MDCA), which is suitable for the fusion of multiple feature sets. The multi-set feature fusion process based on the MDCA technique can be summarized as follows: It is assumed that a total of *m* sets of features, $X_i$, $i = 1,2,...,m$, need to be fused, with their ranks arranged in descending order, i.e. $rank(X_1) \geq rank(X_2) \geq ... \geq rank(X_m)$. When implementing the MDCA-based fusion, the DCA is successively applied to only two sets of features at one time: *the two feature sets with the highest ranks are firstly fused together, and the fusion result will be fused with the feature set of the next highest rank, and so on. If the feature sets of the identical rank exist, they can be fused together at any time* [13].

For the binary classification of image interestingness, different interestingness cues mentioned above are related to multiple types of features respectively. For each feature type, different feature representation methods are harnessed to generate its feature sets. In our work, the DCA (if the total number of feature sets is 2) or the MDCA (if the total number of feature sets is larger than 2) technique was used to implement feature fusion for different feature types belonging to individual cues, to obtain one feature vector of more powerful discriminant ability. For each of the three cues above, the feature vectors associated with different feature types are then concatenated into one extended feature vector to comprehensively describe them respectively.

**3.4　SimpleMKL-based binary classification of image interestingness**

In the existing research work, after obtaining the features used to describe the interestingness of images, SVM is often employed to implement the binary interestingness classification [5-10]. More specially, a single kernel function is typically used to implement the mapping of the feature space into the class space, where the mapping function can be implicitly built by a kernel function. Theoretically, kernel-based method is an effective method to solve the problem of nonlinear classification, but, in some cases, the kernel machine composed of a single kernel function cannot meet practical requirements while dealing with heterogeneous data, so it is an inevitable choice to combine in a nonlinear way the stengths of multiple kernel functions to obtain better results.

　　From the above description in Section 3.2, to comprehensively describe the interestingness of an image, three categories of cues are empirically leveraged, each of which is related to a couple of feature types. A a result, the heterogeneous characteristics of data are remarkable, for which one has to resort to the multiple kernel learning (MKL) techniques. MKL is a machine learning methodology based on the convex combination of predefined kernels where the combination of kernels can be determined by using different learning methods, linear or nonlinear [12]. Compared with the use of single kernel, the use of different kernel combinations enables in practice the application of kernel methods more flexible and more interpretable.

　　In this paper, the SimpleMKL algorithm proposed by Rakotomamonjy et al. [14] is adopted to address the multiple kernel learning problem. The algorithm includes the weighting of the candidate kernel functions in the objective function of the standard SVM, tranformation of the objective function into a convex and smooth function and and its optimization. More specially, the main tasks of this algorithm are composed of [14]: (1) Fix the weights of kernel functions where the optimization problem is reduced to the standard SVM optimization; (2) Update the weights according to the gradient descent direction of the objective function. Please refer to [14] for the detailed implementation process.

In general, as the kernelization trick, some kernel function is needed to implicitly embed the features in the low-dimensional space into a high- or indefinite-dimensional space [35], making the original linearly-unseparable data become linearly separable. In our work, the kernel functions involved in the SimpleMKL algorithm include the Gaussian kernel (also called Radial Basis Function kernel) and the polynomial kernel functions. More specially, for the features used to describe the unusualness cue, the Gaussian kernel function is used for the feature mapping:

$$K_{RBF}(\mathbf{u},\mathbf{v}) = \exp\left(-\frac{\|\mathbf{u}-\mathbf{v}\|_2^2}{2\sigma^2}\right) \quad (15)$$

where $\mathbf{u}$ and $\mathbf{v}$ are two feature vectors, and $\sigma$ is the kernel parameter; for the features to describe the cues of aesthetics and general preferences, we employed the polynomial kernel functions of degree 2 and degree 3, respectively, as follows

$$K_P^2(\mathbf{u},\mathbf{v}) = (\mathbf{u}^T\mathbf{v}+1)^2 \quad (16)$$

$$K_P^3(\mathbf{u},\mathbf{v}) = (\mathbf{u}^T\mathbf{v}+1)^3 \quad (17)$$

As a result, three feature spaces can be constructed. The gradient descent method was then adopted to get the weights, $d_1, d_2, d_3$, for these funcitons to produce the combined kernel function as $K = d_1 K_{RBF} + d_2 K_P^2 + d_3 K_P^3$.

It should be noted that, for the heterogeneous features extracted by different methods in Section 3.2, each feature is firstly normalized to have mean of 0 and standard deviation of 1 before its mapping into the high-dimensional feature space, to deal with the issue that there exists more or less difference in terms of the distributions of these features.

### 3.5 Algorithmic implementation

Basically speaking, the implementation process of binary interestingness classification of images can be divided into three phases: the extraction of interestingness features, the training of interestingness classification model with training images and the interestingness classification of test images. For the sake of completeness, the algorithmic pseudo-codes for the training and test phases are outlined in the following.

**Training phase**

**Input:** $(I_i, l_i)_{i=1}^{N}$, where $I_i$ is the *i*th training sample, $l_i$ is the label of the training sample, and $N$ is the number of training samples

**Output:** The binary interestingness classification model $F(\mathbf{X})$

**for** *i*=1:*N*

1) Extract the interestingness features of the *i*th training sample based on the interestingness cues and the relevant feature extraction methods;

**end for**

2) Use the DCA/MDCA technique to obtain the fused feature set $\mathbf{X} = (\mathbf{x}_1, \mathbf{x}_2, ..., \mathbf{x}_N)$;

3) With the feature set $\mathbf{X}$ and the label set $L_{\text{training}} = (l_1, l_2, ..., l_N)$, use the SimpleMKL algorithm to train the binary interestingness classification model $F(\mathbf{X})$.

**Test phase**

**Input:** $(I_j)_{j=1}^{M}$, where $I_j$ is the *j*th test sample, $M$ is the number of test samples, and the binary interestingness classification model $F(\mathbf{X})$

**Output:** The lable set $\hat{L}_{\text{test}} = (\hat{l}_1, \hat{l}_2, ..., \hat{l}_M)$ for the test data

**for** *j*=1:*M*

1) Extract the interestingness features of the *j*th test sample based on the interestingness cues and the relevant feature extraction methods;

**end for**

2) Use the DCA/MDCA technique to obtain the fused feature set $\hat{\mathbf{X}} = (\hat{\mathbf{x}}_1, \hat{\mathbf{x}}_2, ..., \hat{\mathbf{x}}_M)$;

3) Use the binary interestingness classification model $F(\mathbf{X})$ to classify the feature set of the test data to obtain the label set $\hat{L}_{\text{test}} = (\hat{l}_1, \hat{l}_2, ..., \hat{l}_M)$.

Fig. 2　The algorithms of the proposed framework in the training and test phases

## 4 Experimental results and analysis

### 4.1 Data Set

The data set used in this paper is the one provided in the competition of Predicting Multimedia Interestingness Task released in 2016 [17], which is composed of Hollywood movie trailers licensed by Creative Commons. The main purpose of this task is to develop and design the automatic interestingness prediction models for the images from the Hollywood movie trailers as well as the video clips.

The whole data set consists of 78 trailers, where each trailer is segmented into video shots, and the middle frame of each shot is taken as the image data. As such , the entire data set includes 7,396 images. A total of 100 annotators participated in the annotation work of these images. In our work, the whole data set was divided into the training set and the test set according to the ratio of 7:3. Figure 3 show some representative image samples in the data set, where the first two rows are interesting samples and the last two rows uninteresting ones.

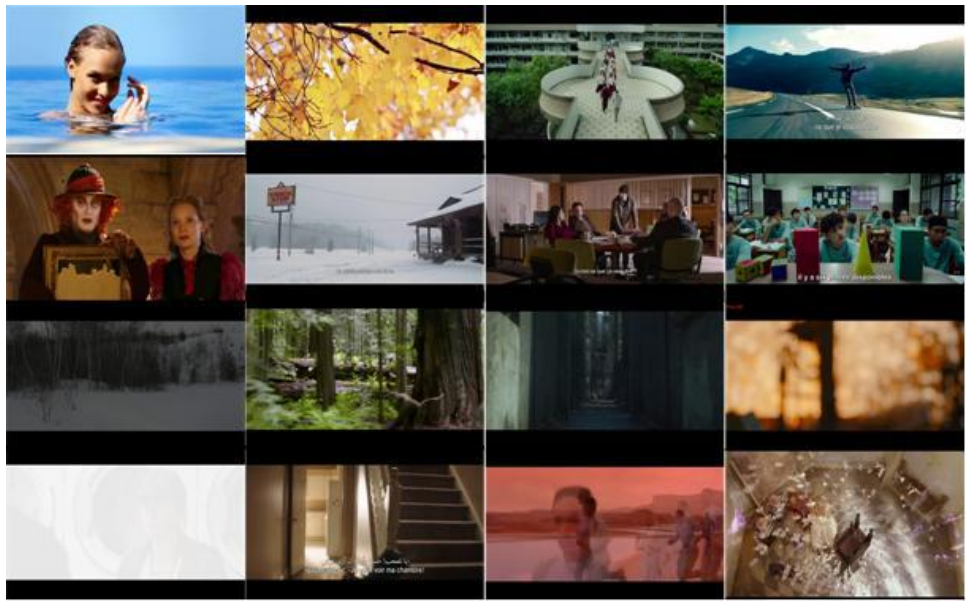

Fig. 3　Some representative samples in the data set
for the competition of Predicting Multimedia Interestingness Task released in 2016

### 4.2 Experimental Settings

In our work, to verify the effectiveness of the proposed framework and its implementation scheme, a total of six sets of experiments have been carried out, which will be described briefly.

In the first set of experiments, the SVM-based binary classification tasks were conducted with the combinations of the interestingness features in [5], [11] and [36] and the combination of interestingness features in this paper.

In the second set of experiments, the SVM-based binary classification tasks were carried out with the concatenated combination of the features used in this paper as well as with the concatenated combination of the fused features by the DCA/MDCA technique, respectively. Additionally, the SimpleMKL-based classification was also completed with the fused features.

The third set of experiments are related to the SVM-based binary classification with the three cues selected in this paper, aiming to investigate separately the contribution of each cue to the interestingness prediction.

In the fourth, the fifth and the sixth sets of experiments, the SVM-based binary classification tasks were performed separately with different types of features across three cues to verify how different types of features contribute to the interestingness predicition.

In all the classification experiments above with the SVM classifier, the optimal kernel function and parameter setting were determined by means of parameter optimization: the polynomial kernel with degree=3 and $\gamma = 32$. For the experiments where the SimpleMKL algorithm was adopted for classification, the combined kernel function obtained from the training data is $K = 0.1126 K_{RBF} + 0.2751 K_P^2 + 0.6123 K_P^3$.

It can be noticed from the six sets of experiments above that, each experiment mainly includes four procedures: (1) feature extraction from the training data and the test data respectively; (2) normalization of the mean and the variance of the obtained features vectors; (3) feature concatenation or feature fusion according to the corresponding rules; and (4) single kernel- or multiple kernels-based interestingness classification.

For the feature concatenations, for each interestingness cue, all types of features involved were simply concatenated; in the same way, the features across the three cues were further concatenated to obtain the final interestingness features. For the feature fusion, the DCA/MDCA-based fusion technique was utilized intensively for the same type of features from each interesting cue, while feature concatenations between different types of features and across different cues were used to form the final feature set.

**4.3 Performance Evaluation**

In our experiments, Accuracy (ACC), Area Under the Curve (AUC) and Receiver Operating Characteristic (ROC) curve were selected as the evaluation indexes of the interestingness binary classification task. The first two are numerical indexes and the last one is the curve-based index, used to comprehensively verify the performance of different interestingness classification schemes.

For a binary classification task, the main purpose is to divide the samples into positive ones or negative ones. Generally speaking, there are four cases with respect to the resutls: True Positive (TP), False Negative (FN), False Positive (FP) and True

Negative (TN), based on which the performance evaluation indexes can be defined as follows.

(1) ACC

ACC is the ratio of correctly classified samples to the total samples and expressed as:

$$ACC = \frac{TP+TN}{TP+TN+FP+FN} \tag{18}$$

(2) ROC curve

The abscissa of the ROC curve is False Positive Rate (FPR), i.e., the proportion of negative samples that are predicted to be positive in all negative samples. The ordinate is the True Positive Rate (TPR), i.e., the proportion of positive samples that are predicted to be positive in all positive samples.

$$FPR = \frac{FP}{TN+FP}, \quad TPR = \frac{TP}{TP+FN} \tag{19}$$

(3) AUC

The area under the ROC curve is the AUC, taking the value of between 0 and 1. Compared with the ROC curve, this index can be used to more intuitively assess the performances of different schemes. The larger the AUC value is, the better the prediction ability of the model.

**4.4 Experimental Results**

**(1) Effects of different combinations on the prediction performance of image interestingness**

In [11], the description of visual interestingness was made by color histogram, scene descriptor and LBP, and the classificaiton task is implemented with the SVM (F1 method). In [36], the SIFT and the scene descriptor Gist were combined to describe visual interestingness, and the classifier was also based on the SVM (F2 method). In [5], a set of features were used to describe visual interestingness, including local outlier coefficient, familiarity, arousal, the compression rate from the original image, the Gist and the SIFT, and the SVM-based classifier was employed to complete the classification task (F3 method). In this paper, as elaborated in Section 3.2, a series of features were extracted to describe the image interestingness in terms of three representative cues, i.e. unusualness, aesthetics and general preferences, and the SVM was still used for interestingness classification (F4 method). Figure 4 and Table 1 show the ROC curves of and the corresponding ACC and AUC values from these four methods, respectively.

Table 1  Quantitative evaluation results with different methods

| | Evaluation Index | ACC | AUC |
|---|---|---|---|
| Methods | F1[11] | 0.885 | 0.952 |
| | F2[36] | 0.883 | 0.953 |
| | F3[5] | 0.892 | 0.975 |
| | F4 | **0.898** | **0.986** |

It can be noted remarkbly from Fig. 4 and Table 1 that, compared to the methods of F1 and F2, F3 and F4 methods improve the performance by introducing more features to describe interestingness cues, especially for the F4 method where the texture, shape and local features are addded and the edge detection and significance map technologies are employed to obtain higher-level complexity features, in place of the complexity features based on the original image as in the F3 method, achieving the highest ACC and AUC values of 0.898 and 0.986, respectively. Therefore, the combination of interestingness features selected in this paper is more comprehensive and more effective in describing image interestingness than those used in [5,11,36].

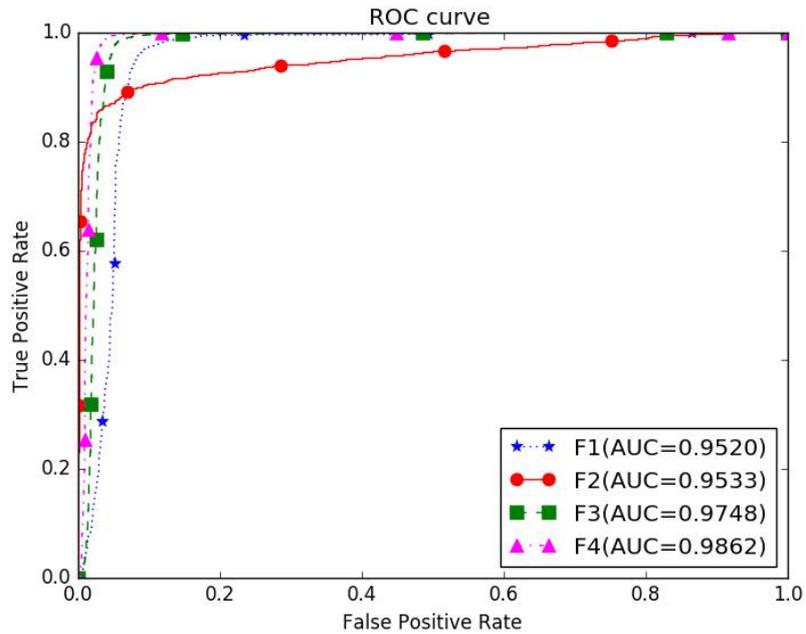

Fig. 4  ROC curves with different methods

## (2) The influence of feature fusion on the prediction performance of image interestingness

In the F4 method, all the extracted features are concatenated directly. To verify the effectiveness of the proposed framework of image interestingness, two different strategies are given. On the one hand, for different types of features, the DCA or MDCA technique is used for the fusion of feature sets belonging to the same type before the concatenation of different types of features to describe the the cues of unusualness, aesthetics and general preferences respectively. Then, the fused and concatenated features for different cues are further concatenated as the input to the SVM classifier (F5 method). On the other hand, after the internal fusion of different types of features as in the F5 mehod, different types of features are concatenated to describe the above three cues respectively, and the SimpleMKL algorithm is then used for classification (F6 method). Figure 5 and Table 2 show the ROC curves of and the corresponding ACC and AUC values from the three methods, respectively.

Table 2   Quantitative evaluation results with different methods

| | Evaluation Index | ACC | AUC |
|---|---|---|---|
| | F4 | 0.898 | 0.986 |
| Methods | F5 | 0.913 | 0.989 |
| | F6 | **0.944** | **0.999** |

According to Fig. 5, the critical points of the ROC curves from the F4, F5 and F6 methods are close to the upper left corner. It can be seen from Table 2, for the ACC index, the value from the F5 method is 0.015 higher than that from the F4 method, and the AUC value is also increased. In addition, the fused feature dimension is reduced from 1,361 dimensions to 257 dimensions, greatly reducing the computational complexity. For the F6 method, the quantitative results are significantly improved than the F5 method, up to 0.944 and 0.999 for the ACC and the AUC indexes respectively, demonstrating together the efficacy of the DCA/MDCA based feature fusion and the multiple-kernel learing strategies.

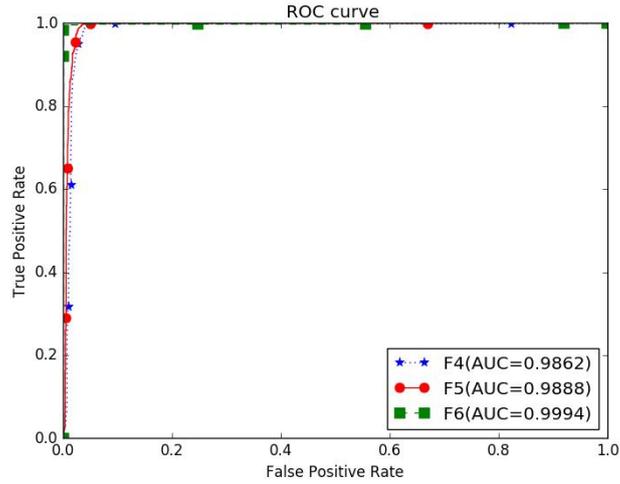

Fig. 5 ROC curves with different strategies

**(3)The effect of different cues on the prediction performance of image interestingness**

Three interestingness cues were utilized in this paper, i.e., unusualness, aesthetics and general preferences, which are denoted as cue1, cue2 and cue3 respectively. This experiment investigates how each of the three cues separately contributes to the prediction performance.

Figure 6 and Table 3 show the ROC curves of and the corresponding ACC and AUC values from the three cues, respectively.

Table 3 Quantitative evaluation results with different cues

| Evaluation Index | | ACC | AUC |
| --- | --- | --- | --- |
| | cue1 | 0.535 | 0.559 |
| Cues | cue2 | 0.861 | 0.909 |
| | cue3 | 0.890 | 0.959 |

According to Fig. 6 and Table 3, among the three cues to describe interestingness, aesthetics and general preferences exhibit good classification accuracy in predicting image interestingness: AUC values are above 0.90, and ACC values are above 0.85 while the values of AUC and ACC from the only use of unusualness cue are muc lower. These results suggest that images that conform to aesthetic rule and general preferences are more likely to be considered interesting, which is the reason why we have paid more attention to the relevant features in describing the cues of aesthetics and general preferences.

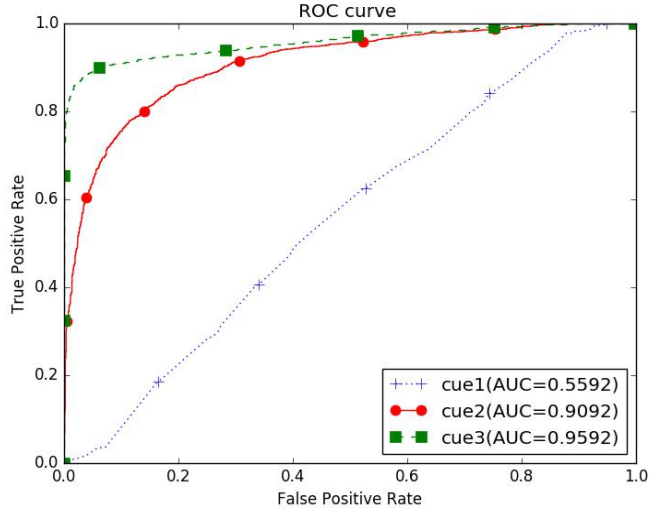

Fig. 6　ROC curves with different interestingness cues

**(4) The influence of different types of features on the prediction of image interestingness**

Based on the experimental results above, we analyzed the contribution of different types of features across unusualness, aesthetics and general preference cues to the prediction of image interestingness.

Figure 7 and Table 4 show the ROC curves as well as the corresponding ACC and AUC values respectively for two different types of features to describe the unusualness cue, i.e., LOF and familiarity.

Table 4　Quantitative evaluation results with different feature types for the unusualness cue

| Evaluation Index | | ACC | AUC |
|---|---|---|---|
| Feature Types | LOF | 0.483 | 0.481 |
| | familiarity | 0.559 | 0.566 |

It can be seen from Table 4 that the unusualness cue has the least contribution to the interestingness prediction among the three selected interestingness cues, where the LOF feature has lower positive effect since the resulting ACC and AUC values are not more than 0.5. In contrast, the familiarity features contributes more in capturing the interestingness of images. For the ROC characteristics in Fig. 7, the familarity feature also brings about better results.

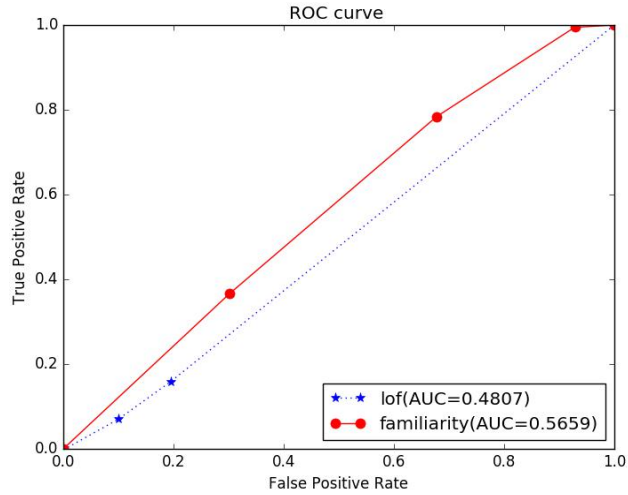

Fig. 7　ROC curves with different feature types for the unusualness cue

Figure 8 and Table 5 show respectively the ROC curves and the corresponding ACC and AUC values by individual use of the five different types of features (arousal, color, texture, complexity, and shape) for describing the aesthetics cue.

Table 5　Quantitative evaluation results with different feature types for the aesthetics cue

| Evaluation Index | | ACC | AUC |
| --- | --- | --- | --- |
| Feature Types | arousal | 0.584 | 0.603 |
| | color | 0.694 | 0.756 |
| | texture | 0.684 | 0.750 |
| | complexity | 0.703 | 0.585 |
| | shape | 0.681 | 0.539 |

Taking into account the results in Figs. 8 and 6 and those in Tables 5 and 3, one can note that it is effective to combine the five types of features above to capture the aesthetic cue. The highest AUC value is obtained solely by the color feature (0.756), the highest ACC value is due to the complexity feature (0.703), and the AUC and ACC values from the combined use of the five features are increased by 0.158 and 0.153, respectively.

In terms of the ACC index, the complexity and color features are more effective in the representation of inage interestingness, indicating that the features extracted from the perspective of the visual perception mechanism of human beings could capture more interestingness content. As far as the AUC is concerned, the color and texture features could better highlight general interestingness of images.

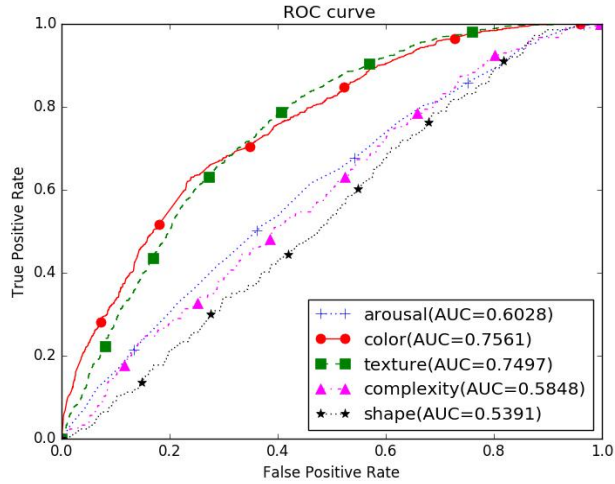

Fig. 8　ROC curves with different feature types for the aesthetics cue

　　Figure 9 and Table 6 show the ROC curves with three different types of features (SIFT, HOG and Gist) for the general preferences cue as well as the corresponding ACC and AUC values respectively.

Table 6　Quantitative evaluation results with different feature types for the general preferences cue

| Evaluation Index | | ACC | AUC |
| --- | --- | --- | --- |
| Feature Types | SIFT | 0.772 | 0.887 |
| | HOG | 0.765 | 0.940 |
| | Gist | 0.706 | 0.773 |

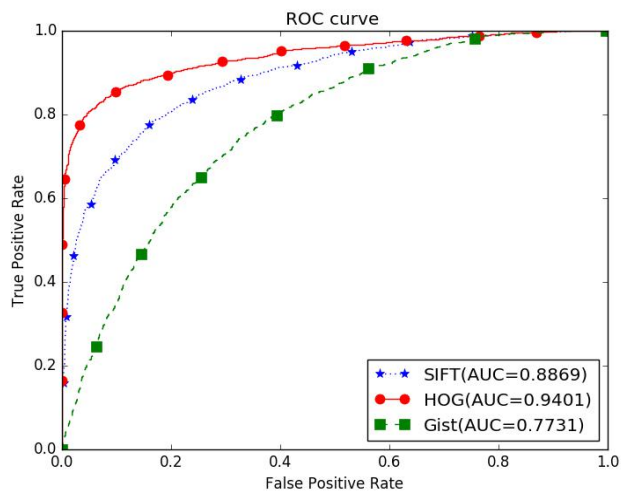

Fig. 9　ROC curves with different feature types for the general preferences cue

Combining the results in Figs. 9 and 6 and those in Tables 6 and 3, it can be seen that the direct concatenation of three types of features above to represent the general preferences cue improves the interestingness prediction performance. More specially, the highest AUC value (0.940) is derived from the sole use of the HOG feature, and the highest ACC value (0.772) is attributed to the use of the SIFT feature, while the combined use of these three features increases the AUC and ACC values by 0.019 and 0.118 respectively.

## 5  Conclusion

In this paper, we have proposed a general binary classification framework to predict the visual interestingness of images, which consists of the extraction of different types of features for the visual interestingness cues of unusualness, aesthetics and general preferences, the discriminant correlation analysis (DCA)-based or the multiset discriminant correlation analysis (MDCA)-based fusion of feature sets belonging to the same type, and the multiple kernel classification based on the simple multiple kernel learning (SimpleMKL) algorithm.

Under the proposed framework, we have addressed two key problems: (1) to reduce feature redundancy in capturing the interestingness cues of images, the DCA or MDCA technique was adopted to fuse multiple feature sets of the same type for individual cues by taking into account the class structure among the samples involved, to describe the three cues above with three sets of compact and representative features; (2) to make good use of the heterogeneity from the three sets of high-level features for these interestingness cues, the SimpleMKL method was employed to enhance the generalization ability of the built model for the task of the binary interestingness classification.

Experiments were conducted on the publicly-released data set provided in the competition of Predicting Multimedia Interestingness Task in 2016. Experimental results from several groups of comparative studies across different interestingness feature combinations, different interestingness cues, as well as different feature types for the three interestingness cues have demonstrated the rationality and effectiveness of the proposed framework in the prediction of image interestingness.

The proposed image interestingness prediction framework is unified and flexible since more cues and more features for each cue could be integrated into it to enhance the binary classification performance with respect to a variety of data sets to meet the requirements in different practical applications. In addition, multi-class interestingness prediction tasks could also be realized by the generalization of the framework to multi-classifier systems.

**Acknowledgements** This study was supported in part by the National Natural Science Foundation of China (No. 31671002), the Scientific Research Foundation of Shaanxi Province for Returned Chinese Scholars (No. 2017004), the Shaanxi Natural Science Foundation (No. 2016JM6046 and No. 2016JM6020), and in part by the Science and Technology Programme Project of Yulin City (No. 2016-20-1) via the Yulin Key Laboratory of Environmental Perception and Information Processing).

**Author contributions** Proposed the idea of the work: Q. Sun, L. Wang; Conceived and designed the experiments: Q. Sun, L. Wang; Performed the experiments: L. Wang; Analyzed data: Q. Sun, L. Wang, L. Zhang, Y. Yang; Wrote and edited manuscript: Q. Sun, L. Wang, M. Li, L.Zhang, Y. Yang.

**Compliance with ethical standards**

**Competing interests** The authors declare that they have no competing interests.